# How does the Mind store Information?

Rina Panigrahy
rinap@google.com
10/1/2019

## Abstract

How we store information in our mind has been a major intriguing open question. We approach this question not from a physiological standpoint as to how information is physically stored in the brain, but from a conceptual and algorithm standpoint as to the right data structures to be used to organize and index information. Here we propose a memory architecture directly based on the *recursive sketching* ideas from [this paper](#) to store information in memory as concise *sketches.* We also give a high level, informal exposition of the recursive sketching idea from the paper that makes use of subspace embeddings to capture deep network computations into a concise sketch. These sketches form an implicit knowledge graph that can be used to find related information via sketches from the past while processing an event.

## Introduction

Say you just returned from a meeting -- how would your mind remember and *organize* all the information related to the meeting? Who were the key members of the meeting, what was spoken, what was in the presentations, roughly how many people were there, what was the room like, what were the details? Such information needs to be organized and indexed in a way so that it can be quickly accessed in future if say I met someone from the meeting later to chat about some related topic. We would like to look for a method to store such information in a way that makes sense computationally -- not necessarily trying to look for exactly how it is done in the brain.

We propose that information related to such events and inputs is stored as a *sketch* as described in [this paper](#) --  the sketch is a concise summary of the multimodal sensory input corresponding to the event.  Since we can often recall  and re-visualize the meeting room, the sketch should be approximately reversible; that is, one can approximately reconstruct the original input and/or its crucial properties, summaries and statistics. The sketch needs to be gracefully *erasable* so that even if a long time has passed and I may have forgotten several details, the high-level properties and statistics of the event may still be preserved in the sketch. These sketches need to be organized so as to reveal a knowledge graph so that I may quickly be able to retrieve information events related to a person from the past when I meet the person again.

The proposed sketching mechanism is based on random subspace embedding that seamlessly captures the coarse and fine attributes of (heterogeneous types) and details of an event with summarizing it as a plain vector in euclidean space. This simple sketch is sufficient to approximately reconstruct the original input and its basic statistics upto some level of accuracy. The sketching mechanism implicitly enables different high level object oriented abstractions such as classes, attributes, references, type-information, modules without explicitly incorporating such ideas into the mechanism operations. Thus it gives a higher level of interpretability and abstraction to the seemingly plain high dimensional sketch vectors. Note that while this certainly may not exactly be how we store information, it could be a useful way of "conceptually" capturing how information may be stored by machine learning systems.

We note that another idea for modeling memory is that of [Neural Turing Machines](#) that attaches memory to a deep network by making memory accesses differentiable. Sketching is a rich field of study that dates back to [the foundational work of Alon, Matias, and Szegedy](#), which can enable neural networks to efficiently summarize information about their inputs. Such methods have led to a [variety of efficient algorithms](#) for basic tasks on massive datasets, such as estimating fundamental statistics (e.g., histogram, [quantiles](#) and [interquartile range](#)), finding popular items (known as [frequent elements](#)), as well as estimating the number of distinct elements (known as *[support size](#)*) and the related tasks of [norms](#) and [entropy estimation.](#)

## Method: Recursive subspace embedding

We will assume that there is a deep network that processes the visual and auditory inputs and outputs higher level concepts such as people, words, topics. We will also assume for simplicity that the network that is modular like the one below. Indeed this a major assumption as the networks learned today are far from having a modular interpretation -- perhaps modules can be interpreted by finding sparse information cuts or by treating each layer as a separate module; nonetheless, we will make this modularity assumption. Each module spits out a vector. Usually while there may be several modules, for a given input, only a few (spare) modules may fire.

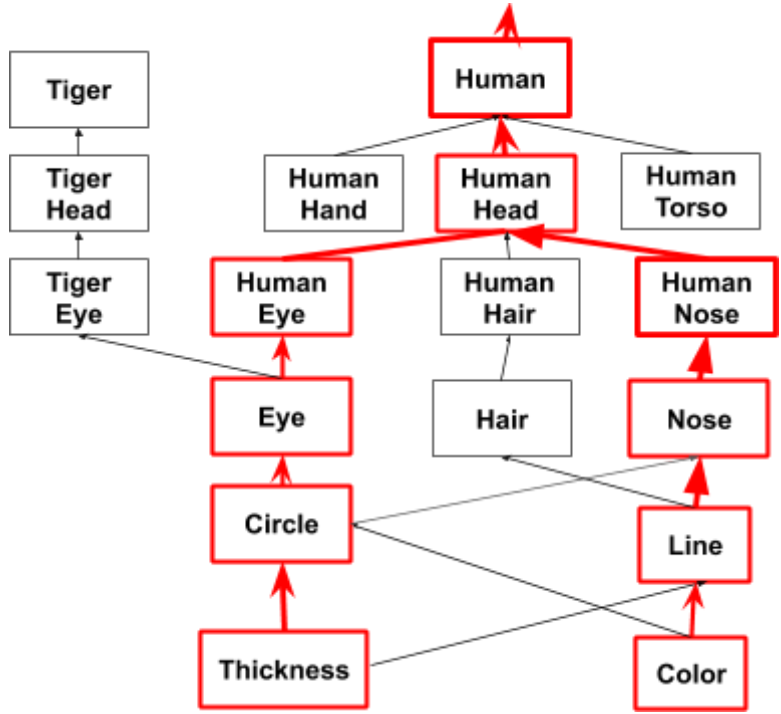

**For simplicity we consider a modular deep network, where each module outputs a vector $x_i$.**

**Subspace Embedding**: Next we note that if there is a bunch of vectors $x_1,\ldots,x_N$ at a given layer output by modules $M_1,\ldots,M_N$ they could be combined into a single sketch vector y that *encodes* which modules fired and with what values. As long as the number of firing modules is sufficiently sparse and the sparsity is sufficiently smaller than the sketch dimension. This idea can be extended recursively to sketch a hierarchy of module firings.

$$y = R_1 x_1 + R_2 x_2 + \ldots + R_N x_N$$

with $M_1, M_1, \ldots, M_N$ below.

**Subspace embedding to sketch the outputs at a given layer. Do this recursively to sketch across layers.**

**Connection to Sparse coding**: The main idea is essentially to use sparse coding. Say you want to encode a sparse vector of N bits $(x_1,\ldots,x_N)$ that is at most k-sparse. A simple way to do this is to encode them by using a collection $r_1,\ldots,r_N$ of random vectors of d-dimensions to map the vector x into a single vector $y = r_1 x_1 + r_2 x_2 + \ldots + r_N x_N$. This encoding y can be thought of as a *sketch* of this sparse vector x. As long as the sparsity k of x is much smaller than sketch dimension d, the vector x can be recovered from the sketch y via sparse-recovery. In our case $x_1,\ldots,x_N$ are not necessarily bits but vectors themselves output by the modules. Again we assume that a small number -- at most k -- modules fire (and the remaining are zero). Again if the sketch dimension d is sufficiently larger than the product of the sparsity k of the set of vectors $x_1,\ldots,x_N$ and the dimensionality of each of these vectors then we can recover the set of vectors.

**Connection to Dictionary Learning:** For the recovery mentioned above we need to have the knowledge of the random matrices $R_1,\ldots R_N$. However note that if we have enough sketch output vectors y over different inputs ($x_1,\ldots,x_N$) then we can use dictionary learning to recover both the set of matrices and the collection of input vectors. To see this think back to the sparse coding of bits using $y = r_1 x_1 + \ldots + r_N x_N$: note that given sufficient outputs over random sparse input vectors x, we learn the coding vectors $r_i$'s as a "dictionary"

**Recursive Sketching:** We generalize the above idea to produce a hierarchical sketch over many layers: The main idea is that to produce the final sketch y, instead of using the outputs $x_i$ of each module we use a recursive sketch of that output $x_i$ of module $M_i$ by combining $x_i$ with the inputs of $M_i$ recursively. Also instead of using random matrices $R_i$, we use matrices from a slightly different distribution. For full details see [here](here) and this [poster](poster)

**Sketch Embeddings**: This produces an embedding in sketch space of every entity/event that we may be processed. Related entities have similar embeddings. These sketches may be organized into a near neighbor search data structure (such as Locality Sensitive Hashing) so that given a sketch one may retrieve related sketches easily. Thus if one meets a person, that person's sketch may be used to retrieve previous events where one may have met the person and from there possibly the contents of the previous discussions with the person.

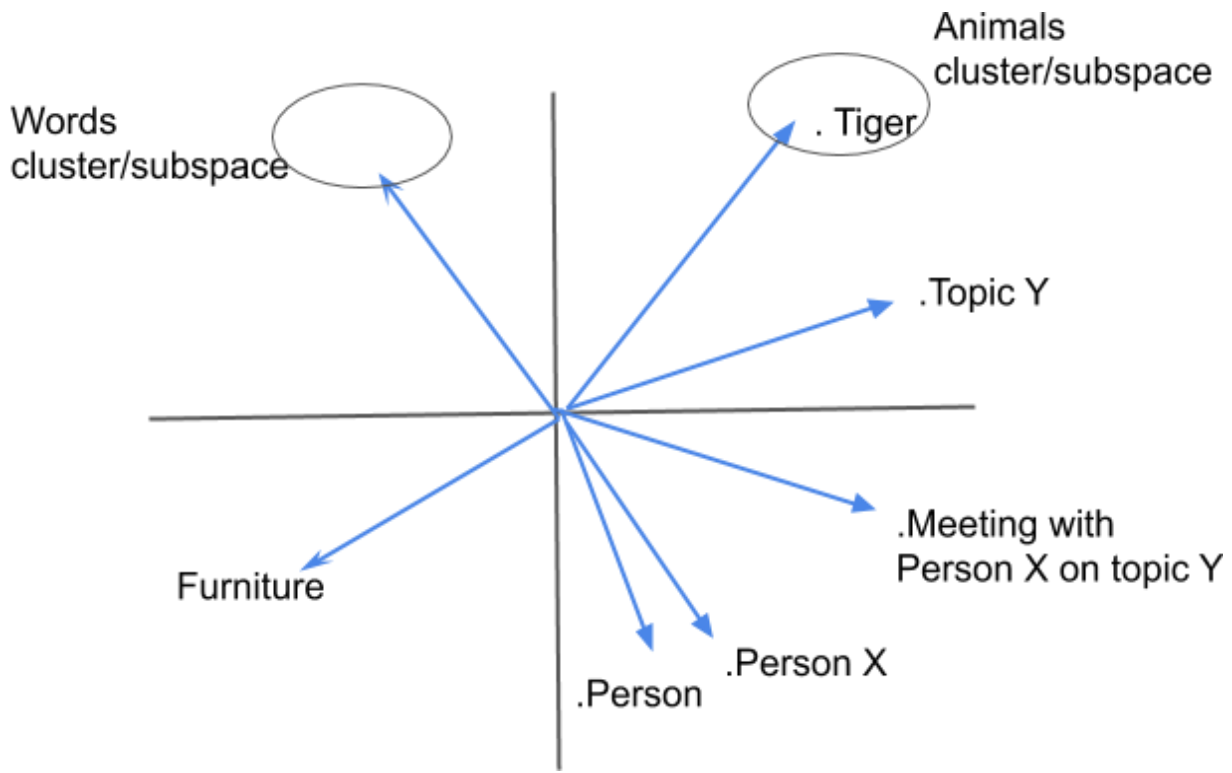

**Different types can be encoded in different directions/subspaces/clusters giving subclusters for subtypes. A new cluster indicates the possibility of a new concept to be learnt**

# Benefits

This sketching / modular network view helps bridge the gap between a non-interpretable homogenous deep network and an object oriented high level programming language view of learning.

# Knowledge Graph

Since related objects have similar sketches (see Theorems in section 3 here), we obtain an implicit *knowledge graph* by placing an edge between a pair of similar sketches. Thus a sketch can be used to retrieve related sketches and component attributes and objects. So if we meet someone then the sketch of that meeting can be used to retrieve sketches of past meetings with that person or other meetings on related topics. These additional sketches can be fed into the current inference process to make a more informed decision/prediction.

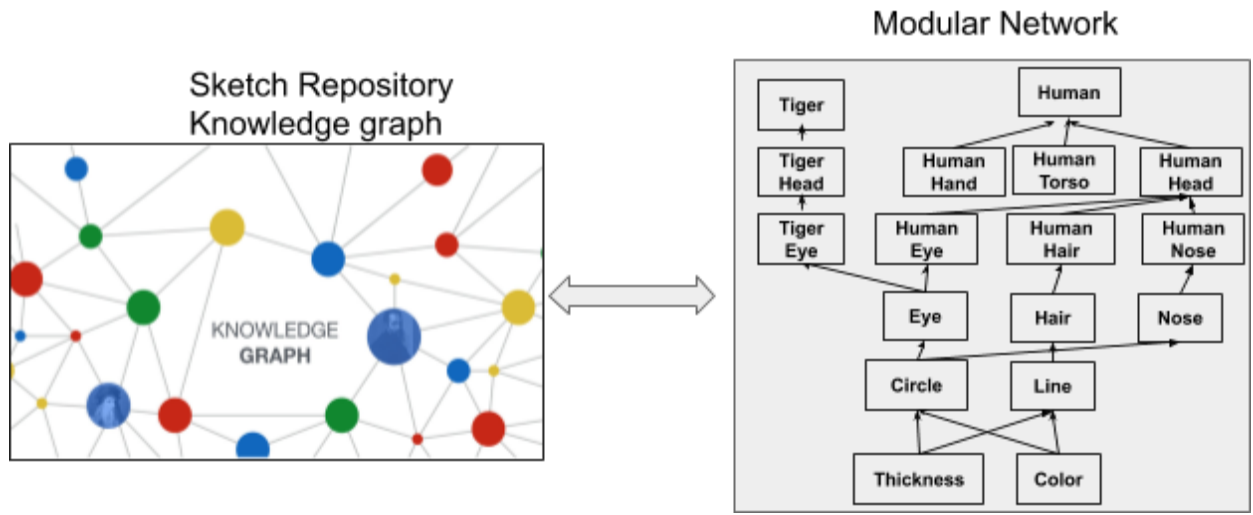

**Since related objects have similar sketches, they have an implicit edge in a knowledge graph. Sketch memory and Modular Network work and inform each other**

**Other Benefits:**

1. **Interpretability**
Since sketches contain not only the final output but also encoding of the incoming inputs and their corresponding modules, they are a more interpretable representation of the computation of the deep network

2. **Attribute Recovery**
From the final sketch the attributes of the objects and the component objects can be recovered upto some level of depth and sparsity.

3. **Graceful erasure**
The information stored in the sketch degrades gracefully as the sketch is forgotten slowly, that is, as bits may be deleted from the sketch.

4. **Simple statistics**
If the sketch is condensing a large number of objects, one can still recover basic statistical operations such as computing histogram / counting / mean /variance.

5. **Similarity preserving**
Sketches of inputs that are *semantically* similar, will result in similar sketches. This could be useful to find past sketches containing the same objects as in the current events sketch. The sketch repository implicitly produces a knowledge graph of sketches

6. **Type representation**
Since type information is implicit in subspace the sketch is present in. Thus the sketching method gives a protocol independent communication method across modules without requiring type support from a high level programming language compiler.

7. **Sketches function as reference pointers**
Sketches can be viewed as a hash of the complex input and can be used as implicit references/pointers to other related sketches and the modules that produced the sketch. The ability to use sketches as pointers allows retrieval, pointers to more detailed sketches, and other objects.

8. **Helps detect a new concept/cluster**
Similar sketches are closer in sketch space, this can be used to identify new concepts by looking for an evolving cluster of sketches not yet associated with the module. For example, if someone has just seen a few tigers, the sketches of these tigers will form a cluster that can be used to create a new 'tiger' module.

# Conclusion

We argued that the recursive sketching ideas from [this paper](#) can be used to create a memory architecture that stores information as a collection of sketches. These sketches can be used to retrieve information stored from past events when related events happen in the future. These sketches also enhance the interpretability and modularity of the deep network computations.

# Acknowledgements

The proposal here is based directly on the ideas from the earlier paper that was joint work with Badih Ghazi and Joshua Wang. The intention here is merely to provide a simpler exposition and propose that it can be used to create a sketch based memory architecture for the mind.